\documentclass[english]{llncs}
\usepackage[T1]{fontenc}
\usepackage[latin1]{inputenc} 
\usepackage{times}
\usepackage{graphicx}
\usepackage{epsfig}
\usepackage{epic}  
\usepackage{float}
\usepackage{color}

\usepackage{gastex}
\usepackage{pspictpg}

\usepackage{amsmath}    
\usepackage{amssymb}    

\makeatletter
\usepackage{babel}
\makeatother

\begin{document}

\vfill{}
\title{Learning rational stochastic  languages}
\vfill{}

\author{François Denis, Yann Esposito, Amaury Habrard}

\institute{Laboratoire d'Informatique Fondamentale de Marseille (L.I.F.) UMR
CNRS 6166
\email{\{fdenis,esposito,habrard\}@cmi.univ-mrs.fr}}

\maketitle

\begin{abstract}
  Given a finite set of words $w_1, \ldots, w_n$ independently drawn
  according to a fixed unknown distribution law $P$ called a
  \emph{stochastic language}, an usual goal in Grammatical Inference
  is to infer an estimate of $P$ in some class of probabilistic
  models, such as \emph{Probabilistic Automata} (PA). Here, we study
  the class ${\cal S}_{{\mathbb R}}^{rat}(\Sigma)$ of
  \emph{rational stochastic languages}, which consists in stochastic
  languages that can be generated by \emph{Multiplicity Automata} (MA)
  and which strictly includes the class
  of stochastic languages generated by PA.  Rational stochastic
  languages have minimal normal representation which may be very
  concise, and whose parameters can be efficiently estimated from
  stochastic samples. We design an efficient inference algorithm DEES
  which aims at building a minimal normal representation of the
  target. Despite the fact that no recursively enumerable class of MA
  computes exactly ${\cal S}_{\mathbb
    Q}^{rat}(\Sigma)$, we show that DEES strongly identifies ${\cal S}_{\mathbb
    Q}^{rat}(\Sigma)$ in the limit. We study the intermediary MA output
  by DEES and show that they compute rational series which converge
  absolutely to one and which can be used to provide stochastic
  languages which closely estimate the target.
\end{abstract}

\section{Introduction}
In probabilistic grammatical inference, it is supposed that data arise
in the form of a finite set of words $w_1, \ldots, w_n$, built on a
predefinite alphabet $\Sigma$, and independently drawn according to a
fixed unknown distribution law on $\Sigma^*$ called a \emph{stochastic
  language}. Then, an usual goal is to try to infer an estimate of
this distribution law in some class of probabilistic models, such as
\emph{Probabilistic Automata} (PA), which have the same expressivity
as Hidden Markov Models (HMM). PA are identifiable in the
limit~\cite{DenisEsposito04}. However, to our knowledge, there exists
no efficient inference algorithm able to deal with the whole class of
stochastic languages that can be generated from PA. Most of the previous
works use restricted subclasses of PA such as Probabilistic
Deterministic Automata
(PDA)~\cite{CarrascoOncina94,ThollardDupontHiguera00}. In the other
hand, Probabilistic Automata are particular cases of
\emph{Multiplicity Automata}, and stochastic languages which can be
generated by multiplicity automata are special cases of \emph{rational
  languages} that we call \emph{rational stochastic languages}.  MA
have been used in grammatical inference in a variant of the exact
learning model of Angluin
\cite{BergadanoVarricchio94,BeimelBergadanoBshoutyKushilevitzVarricchio96,BeimelBergadanoBshoutyKushilevitzarricchio00}
but not in probabilistic grammatical inference. Let us design by
${\cal S}_K^{rat}(\Sigma)$, the class of rational stochastic languages
over the semiring $K$.  When $K={\mathbb Q}^+$ or $K={\mathbb R}^+$,
${\cal S}_K^{rat}(\Sigma)$ is exactly the class of stochastic
languages generated by PA with parameters in $K$.  But, when
$K={\mathbb Q}$ or $K={\mathbb R}$, we obtain strictly greater classes
which provide several advantages and at least one drawback:
elements of ${\cal S}_{K^+}^{rat}(\Sigma)$ may have significantly
smaller representation in ${\cal S}_{K}^{rat}(\Sigma)$ which is
clearly an advantage from a learning perspective; elements of ${\cal
  S}_{K}^{rat}(\Sigma)$ have a minimal normal representation while
such normal representations do not exist for PA; parameters of these
minimal representations are directly related to probabilities of some
natural events of the form $u\Sigma^*$, which can be efficiently
estimated from stochastic samples; lastly, when $K$ is a field,
rational series over $K$ form a vector space and efficient linear
algebra techniques can be used to deal with rational stochastic
languages. However, the class ${\cal S}_{{\mathbb Q}}^{rat}(\Sigma)$
presents a serious drawback : there exists no recursively enumerable
subset of MA which exactly generates it~\cite{DenisEsposito04}.
Moreover, this class of representations is unstable: arbitrarily close
to an MA which generates a stochastic language, we may find MA whose
associated rational series $r$ takes negative values and is not
absolutely convergent: the global weight $\sum_{w\in \Sigma^*}r(w)$
may be unbounded or not (absolutely) defined.  However, we show that
${\cal S}_{\mathbb Q}^{rat}(\Sigma)$ is strongly identifiable in the
limit: we design an algorithm DEES such that, for any target
$P\in{\cal S}_{\mathbb Q}^{rat}(\Sigma)$ and given access to an
infinite sample $S$ drawn according to $P$, will converge in a finite
but unbounded number of steps to a minimal normal representation of
$P$. Moreover, DEES is efficient: it runs
within polynomial time in the size of the input and it computes a
minimal number of parameters with classical statistical rates of
convergence. However, before converging to the target, DEES output MA
which are close to the target but which do not compute stochastic
languages. The question is: what kind of guarantees do we have on these
intermediary hypotheses and how can we use them for a probabilistic
inference purpose? We show that, since the algorithm aims at building
a minimal normal representation of the target, the intermediary
hypotheses $r$ output by DEES have a nice property: they absolutely
converge to 1, i.e. $\overline{r}=\sum_{w\in\Sigma^*}|r(w)|<\infty$
and $\sum_{k\geq 0}r(\Sigma^k)=1$.
As a consequence, $r(X)$ is defined without ambiguity for any
$X\subseteq \Sigma^*$, and it can be shown that
$N_r=\sum_{r(u)<0}|r(u)|$ tends to 0 as the learning proceeds. Given
any such series $r$, we can efficiently compute a stochastic language
$p_r$, which is not rational, but has the property that
$e^{N_r/\overline{r}}\leq p_r(u)/r(u)\leq 1$ for any word $u$ such that
$r(u>0)$. Our conclusion is that, despite the fact that no
recursively enumerable class of MA represents the class of rational
stochastic languages, MA can be used efficiently to infer such
stochastic languages. 

Classical notions on stochastic languages, rational series, and
multiplicity automata are recalled in Section~\ref{preliminaires}. We
study an example which shows that the representation of rational
stochastic languages by MA with real parameters may be very concise. We
introduce our inference algorithm DEES in Section~\ref{idlim} and we show
that ${\cal S}_{\mathbb Q}^{rat}(\Sigma)$ is strongly indentifiable in
the limit.    We study the properties of the MA output
by DEES in Section~\ref{lrsl} and we show that they define absolutely
convergent rational series which can be used to compute stochastic
languages which are estimates of the target.

\section{Preliminaries}\label{preliminaires}
\paragraph{Formal power series and stochastic languages.}
Let $\Sigma^{*}$ be the set of words on the finite alphabet
$\Sigma$. The empty word is denoted by $\varepsilon$ and the length of
a word $u$ is denoted by $|u|$.  For any integer $k$, let
$\Sigma^k=\{u\in \Sigma^{*}: \ |u|=k\}$ and $\Sigma^{\leq k}=\{u\in
\Sigma^{*}: \ |u|\leq k\}$. We denote by $<$ the
length-lexicographic order on $\Sigma^*$. A subset $P$ of $\Sigma^{*}$
is \emph{prefixial} if for any $u,v\in \Sigma^*$, $uv\in P\Rightarrow
u\in P$. For any $S\subseteq  \Sigma^*$, let $pref(S)=\{u\in \Sigma^*:
\exists v\in \Sigma^*, uv\in S\}$ and
$fact(S)=\{v\in \Sigma^*:
\exists u,w\in \Sigma^*, uvw\in S\}$. 

Let $\Sigma$ be a finite alphabet and $K$ a semiring. A \emph{formal
  power series} is a mapping $r$ of $\Sigma^*$ into $K$. In this
paper, we always suppose that $K\in \{{\mathbb R},{\mathbb Q},
{\mathbb R}^+, {\mathbb Q}^+ \}$.  The set of all formal power series
is denoted by $K\langle\langle \Sigma\rangle\rangle$. Let us denote by
$supp(r)$ the \emph{support} of $r$, i.e. the set $\{w\in \Sigma^*:
r(w)\neq 0\}$.

A \emph{stochastic language} is a formal series $p$ which takes its
values in ${\mathbb R}^+$ and such that $\sum_{w\in \Sigma^*}p(w)=1$.
For any language $L\subseteq
\Sigma^*$, let us denote $\sum_{w\in L}p(w)$ by $p(L)$.  The set of all
stochastic languages over $\Sigma$ is denoted by ${\cal S}(\Sigma)$.
For any stochastic language $p$ and any word $u$ such that
$p(u\Sigma^*)\neq 0$, we define the stochastic language $u^{-1}p$ by
$u^{-1}p(w)=\frac{p(uw)}{p(u\Sigma^*)}\cdot$
$u^{-1}p$ is called the
\emph{residual language} of $p$ wrt $u$. Let us denote by $res(p)$ the
set $\{u\in \Sigma^*: p(u\Sigma^*)\neq 0\}$ and by
$Res(p)$ the set $\{u^{-1}p: u\in res(p)\}$. We call \emph{sample} any
finite sequence of words. Let $S$ be a sample. We denote by $P_{S}$
the empirical distribution on $\Sigma^*$ associated with $S$. A
\emph{complete presentation} of $P$ is an infinite sequence $S$ of
words independently drawn  according to $P$. We denote by $S_n$ the sequence
composed of the $n$
first words of $S$.
 We shall make a frequent use of the Borel-Cantelli Lemma which states that if $(A_k)_{k\in {\mathbb N}}$ is
a sequence of events such that $\sum_{k\in {\mathbb
N}}Pr(A_k)<\infty$, then the probability that a finite number of $A_k$
occurs is 1.
\paragraph{Automata.}\label{MA}
Let $K$ be a semiring. A $K$-\emph{multiplicity automaton (MA)} is a
  5-tuple $\langle \Sigma,Q,$ $\varphi, \iota,\tau\rangle $ where $Q$
  is a finite set of states, $\varphi:Q\times\Sigma\times Q\rightarrow
  K$ is the transition function, $\iota:Q\rightarrow K$ is the
  initialization function and $\tau:Q\rightarrow K$ is the termination
  function. Let $Q_I=\{q\in Q|\iota(q)\neq 0\}$ be the set of
  \emph{initial states} and $Q_T=\{q\in Q|\tau(q)\neq 0\}$ be the set
  of \emph{terminal states}. The \emph{support} of an MA
  $A=\left\langle \Sigma,Q,\varphi,\iota,\tau\right\rangle$ is the NFA
  $supp(A)=\langle \Sigma,Q,Q_I, Q_T, \delta\rangle$ where
  $\delta(q,x)=\{q'\in Q|\varphi(q,x,q')\neq 0\}$.  We extend
  the transition function $\varphi$ to $Q\times\Sigma^*\times Q$ by
  $\varphi(q,wx,r)=\sum_{s\in Q}\varphi(q,w,s)\varphi(s,x,r)$ and
  $\varphi(q,\varepsilon,r)=1$ if $q=r$ and $0$ otherwise, for any
  $q,r\in Q$, $x\in \Sigma$ and $w\in \Sigma^*$. For any finite subset
  $L\subset \Sigma^*$ and any $R\subseteq Q$, define
  $\varphi(q,L,R)=\sum_{w\in L, r\in R}\varphi(q,w,r)$.

For any MA $A$,
let $r_A$ be the series defined by $r_A(w)=\sum_{q, r\in
Q}\iota(q)\varphi(q,w,r)\tau(r).$ For any $q\in Q$, we define the
series $r_{A,q}$ by $r_{A,q}(w)=\sum_{r\in Q}\varphi(q,w,r)\tau(r).$
A state $q\in Q$ is \emph{accessible}
(resp. \emph{co-accessible}) if there exists $q_0\in Q_I$
(resp. $q_t\in Q_T$) and $u\in \Sigma^*$ such that
$\varphi(q_0,u,q)\neq 0$ (resp. $\varphi(q,u,q_t)\neq 0$).  An MA is
\emph{trimmed} if all its states are accessible and
co-accessible. 
From now, we only consider trimmed MA.

A \emph{Probabilistic Automaton (PA)} is a trimmed MA $\left\langle
\Sigma,Q,\varphi,\iota,\tau\right\rangle $ s.t. $\iota, \varphi$ and $\tau$
take their values in $[0,1]$, such that $\sum_{q\in Q}\iota(q)=1$ and
for any state $q$, $\tau(q)+\varphi(q,\Sigma,Q)=1$. Probabilistic
automata generate stochastic languages.
 A \emph{Probabilistic Deterministic Automaton (PDA)} is a PA whose
support is deterministic. 

\begin{figure}[htbp]
 \epsfig{figure=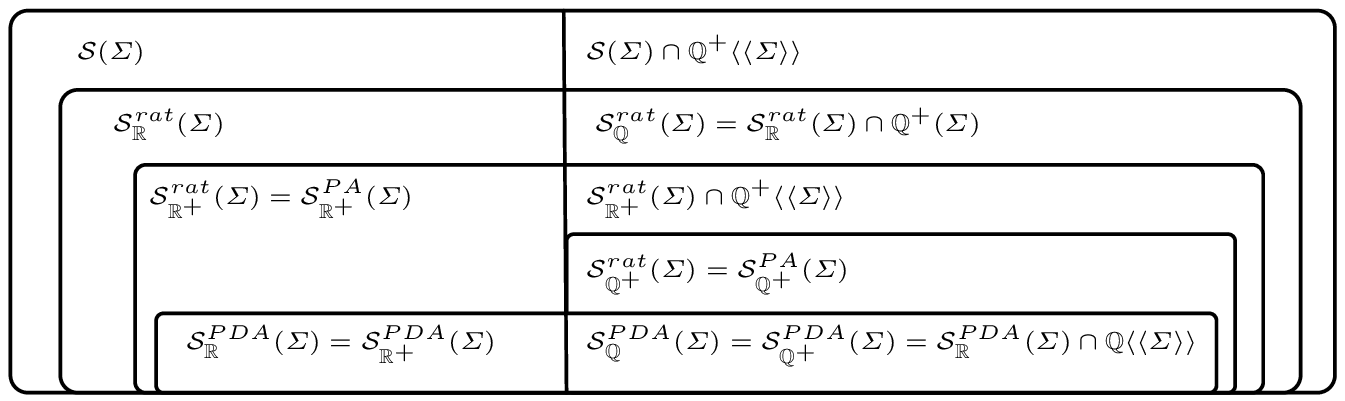, width=12cm}
   \caption{Inclusion relations between classes of rational stochastic
   languages.}
 \label{recapitulatif}
   \end{figure}

For any class $C$ of multiplicity automata over $K$, let us denote by
${\cal S}_K^C(\Sigma)$ the class of all stochastic languages which are
recognized by an element of $C$.
\paragraph{Rational series and rational stochastic languages.}
Rational series have several characterization~(\cite{SalomaaSoittola78,BerstelReutenauer84,Sakarovitch03}). Here, we shall say that a formal power series over
$\Sigma$ is $K$-rational iff there exists a $K$-multiplicity automaton
$A$ such that $r=r_A$, where $K\in \{{\mathbb R}, {\mathbb R^+},
{\mathbb Q}, {\mathbb Q^+}\}$. Let us denote by $K^{rat}\langle\langle
\Sigma\rangle\rangle$ the set of $K$-rational series over $\Sigma$ and
by ${\cal
  S}_K^{rat}(\Sigma)=K^{rat}\langle\langle\Sigma\rangle\rangle\cap
{\cal S}(\Sigma)$, the set of \emph{rational stochastic languages}
over $K$. Rational stochastic languages have been studied in
 \cite{DenisEsposito06} from a language theoretical point of
 view. Inclusion relations between classes of rational stochastic
 languages are summarized on Fig~\ref{recapitulatif}. It is worth
 noting that ${\cal S}^{PDA}_{\mathbb R}(\Sigma)\subsetneq {\cal
 S}^{PA}_{\mathbb R}(\Sigma)\subsetneq {\cal S}^{rat}_{\mathbb
 R}(\Sigma)$.

Let $P$ be a rational stochastic language. The MA $A=\left\langle
\Sigma,Q,\varphi,\iota,\tau\right\rangle $ is a \emph{reduced
representation} of $P$ if (i) $P=P_A$, (ii) $\forall q\in Q,
P_{A,q}\in {\cal S}(\Sigma)$ and (iii) the set $\{P_{A,q}: q\in Q\}$
is linearly independent. It can be shown that $Res(P)$ spans a finite
dimensional vector subspace $[Res(P)]$ of ${\mathbb
R}\langle\langle\Sigma\rangle\rangle$. Let $Q_P$ be the smallest
subset of $res(P)$ s.t. $\{u^{-1}P: u\in Q_P\}$ spans $[Res(P)]$. It is
a finite prefixial subset of $\Sigma^*$. Let $A=\left\langle
\Sigma,Q_P,\varphi,\iota,\tau\right\rangle $ be the MA defined by:
\begin{itemize}
\item $\iota(\varepsilon)=1$, $\iota(u)=0$ otherwise; $\tau(u)=u^{-1}P(\varepsilon)$,
\item $\varphi(u,x,ux)=u^{-1}P(x\Sigma^*)$ if $u,ux\in Q_P$ and $x\in \Sigma$,
\item $\varphi(u,x,v)=\alpha_vu^{-1}P(x\Sigma^*)$ if $x\in \Sigma$, $ux\in (Q_P\Sigma\setminus Q_P) \cap res(P)$ and $(ux)^{-1}P=\sum_{v\in Q_P}\alpha_vv^{-1}P$.
\end{itemize}
It can be shown that $A$ is a reduced representation of $P$; $A$ is
called the \emph{prefixial reduced representation} of $P$. Note that
the parameters of $A$ correspond to natural components of the residual
of $P$ and can be estimated by using samples of $P$. 

We give below an example of a rational stochastic
 language which cannot be generated by a PA. Moreover, for any integer
 $N$ there exists a rational stochastic language which can be generated
 by a multiplicity automaton with 3 states and such that the smallest
 PA which generates it has $N$ states. That is, considering rational
 stochastic language makes it possible to deal with stochastic languages which
 cannot be generated by PA; it also permits to significantly decrease
 the size of their representation.

\begin{figure}[htbph]
          \centering
          \epsfig{figure=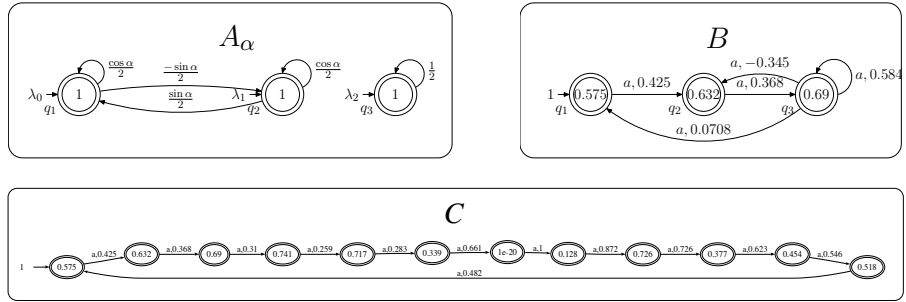, width=12cm}
\caption{When $\lambda_0=\lambda_2=1$ and $\lambda_1=0$, the MA
$A_{\pi/6}$ defines a stochastic language $P$ whose prefixed reduced
representation is the MA $B$ (with approximate values on
transitions). In fact, $P$ can be computed by a PDA and the smallest PA
computing it is $C$.}\label{fig:FMA}
        \end{figure}

 \begin{proposition}For any $\alpha\in {\mathbb R}$, let 
   $A_{\alpha}$ be the MA described on Fig.~\ref{fig:FMA}. Let
   $S_{\alpha}=\{(\lambda_0, \lambda_1, \lambda_2)\in {\mathbb R}^3:
   r_{A_{\alpha}}\in {\cal S}(\Sigma)\}$. If $\alpha/(2\pi) =p/q\in
   {\mathbb Q}$ where $p$ and $q$ are relatively prime, $S_{\alpha}$
   is the convex hull of a polygon with $q$ vertices which are the
   residual languages of any one of them. If $\alpha/(2\pi) \not \in
   {\mathbb Q}$, $S_{\alpha}$ is the convex hull of an ellipse, any
   point of which, is a stochastic language which cannot be computed by
   a PA.
 \end{proposition}

 \begin{proof}[sketch]

   Let $r_{q_0}$, $r_{q_1}$ and $r_{q_2}$ be the series associated
   with the states of $A_{\alpha}$. We have
   $$r_{q_0}(a^n)=\frac{\cos n\alpha-\sin n\alpha}{2^n},
   r_{q_1}(a^n)=\frac{\cos n\alpha+\sin n\alpha}{2^n}\textrm{ and
   }r_{q_2}(a^n)=\frac{1}{2^n}.$$ The sums $\sum_{n\in {\mathbb
   N}}r_{q_0}(a^n), \sum_{n\in {\mathbb N}}r_{q_1}(a^n)$ and
   $\sum_{n\in {\mathbb N}}r_{q_2}(a^n)$ converge since
   $|r_{q_i}(a^n)|=O(2^{-n})$ for $i=0,1,2$. Let us denote
   $\sigma_i=\sum_{n\in {\mathbb N}}r_{q_i}(a^n)$ for $i=0,1,2$. Check
   that
   $$\sigma_0=\frac{4-2\cos{\alpha}-2\sin{\alpha}}{5-4\cos{\alpha}},\
   \sigma_1=\frac{4-2\cos{\alpha}+2\sin{\alpha}}{5-4\cos{\alpha}}\textrm{
   and }\sigma_2=2.$$ 
  
   Consider the 3-dimensional vector subspace ${\cal V}$ of ${\mathbb
     R}\langle\langle \Sigma\rangle\rangle$ generated by $r_{q_0}$,
     $r_{q_1}$ and $r_{q_2}$ and let
     $r=\lambda_0r_{q_0}+\lambda_1r_{q_1}+\lambda_2r_{q_2}$ be a
     generic element of ${\cal V}$.  We have $\sum_{n\in {\mathbb
     N}}r(a^n)=\lambda_0\sigma_0+\lambda_1\sigma_1+\lambda_2\sigma_2$. The
     equation $\lambda_0\sigma_0+\lambda_1\sigma_1+\lambda_2\sigma_2=1$
   defines a plane ${\cal H}$ in ${\cal V}$.

 Consider the constraints $r(a^n)\geq 0$ for any $n\geq 0$. The elements $r$ of
   ${\cal H}$ which satisfies all the constraints $r(a^n)\geq 0$ are
   exactly the
   stochastic languages in ${\cal H}$.





 If $\alpha/(2\pi)=k/h \in {\mathbb Q}$ where $k$ and $h$ are relatively
   prime, the set of constraints $\{r(a^n)\geq 0\}$ is finite: it
   delimites a convex regular polygon $P$ in the plane ${\cal H}$. Let
   $p$ be a vertex of $P$. It can be shown that its residual languages
   are exactly the $h$ vertices of $P$ and any PA generating $p$ must have
   at least $h$ states.

 If $\alpha/(2\pi) \not \in {\mathbb Q}$, the constraints delimite an
   ellipse $E$. Let $p$ be an element of $E$. It can be shown, by using techniques developed in~\cite{DenisEsposito06},  that its
   residual languages are dense in $E$ and that no PA can generate $p$. 
   \qed


 \end{proof}

\paragraph{Matrices.}
We consider the Euclidan norm on ${\mathbb R}^n$: $\Vert(x_1, \ldots,
x_n)\Vert=(x_1^2+\ldots +x_n^2)^{1/2}$. For any $R\geq 0$, let us
denote by $B(0,R)$ the set $\{x\in {\mathbb R}^n: \Vert x\Vert\leq
R\}$. The induced norm on the set of $n\times n$ square matrices $M$
over ${\mathbb R}$ is defined by: $\Vert M\Vert =sup\{\Vert Mx\Vert
:x\in{\mathbb R}^n\textrm{ with }\Vert x\Vert =1 \}.$
Some properties of the induced norm: $\Vert Mx\Vert \leq \Vert M\Vert \cdot \Vert x\Vert $ for all $M\in {\mathbb R}^{n\times
      n}, x\in {\mathbb R}^n$; $\Vert MN\Vert \leq \Vert M\Vert \cdot\Vert N\Vert $ for all $M,N\in {\mathbb R}^{n\times
      n}$; $\lim_{k\rightarrow \infty}\Vert M^k\Vert ^{1/k}=\rho(M)$ where
        $\rho(M)$ is the spectral radius of $M$, i.e. the maximum
        magnitude of the eigen values of $M$ (Gelfand's Formula). 

\section{Identifying ${\cal S}^{rat}_{\mathbb Q}(\Sigma)$ in the limit.}\label{idlim}

Let $S$ be a non empty finite sample of $\Sigma^*$, let $Q$ be
prefixial subset of $pref(S)$, let $v\in pref(S)\setminus Q$, and let
$\epsilon>0$.  We denote by $I(Q, v,S,\epsilon)$ the following set of
inequations over the set of variables $\{x_u|u\in Q\}$: \small{$$I(Q,
v,S,\epsilon)=\{|v^{-1}P_S(w\Sigma^*)-\sum_{u\in
Q}x_uu^{-1}P_S(w\Sigma^*)|\leq \epsilon | w\in
fact(S)\}\cup\{\sum_{u\in Q}x_u=1\}.$$}
\normalsize
Let DEES be the following algorithm: \smallskip \hrule \smallskip 

\noindent \texttt{\textbf{\small Input}}\texttt{\small : a sample $S$}{\small \par}

\noindent \texttt{\textbf{\small 0utput}}\texttt{\small : a prefixial reduced
MA $A=\left\langle \Sigma,Q,\varphi,\iota,\tau\right\rangle $}{\small \par}

\noindent \texttt{\small $Q\leftarrow\left\{ \epsilon\right\} $,
$\iota(\epsilon)=1$, $\tau(\epsilon)=P_S(\epsilon)$,
$F\leftarrow\Sigma\cap pref(S)$}{\small \par}

\noindent \texttt{\small while $F\neq\emptyset$ do \{}{\small \par}

\noindent \texttt{\small ~~~}{\small $v=ux=Min F$ where $u\in \Sigma^*$
and $x\in \Sigma$, $F\leftarrow F\setminus\left\{ v\right\}
$}{\small \par}

\noindent \texttt{\small ~~~if} {\small $I(Q, v,S,|S|^{-1/3})$
has no solution} \texttt{\small then\{}{\small \par}

\noindent \texttt{\small ~~~~~~$Q\leftarrow Q\cup\left\{ v\right\} $,
$\iota(v)=0$, $\tau(v)=P_S(v)/P_S(v\Sigma^*)$,}

\noindent \texttt{\small ~~~~~~
$\varphi(u,x,v)=P_S(v\Sigma^*)/P_S(u\Sigma^*)$,$F\leftarrow
F\cup\{vx\in res(P_S)|x\in\Sigma$\}\}}

\noindent \texttt{\small ~~~}\texttt{\small else\{}{\small \par}

\noindent \texttt{\small ~~~~~~let $(\alpha_w)_{w\in Q}$ be a solution of $I(Q, v,S,|S|^{-1/3})$}

\noindent \texttt{\small ~~~~~~$\varphi(u,x,w)=\alpha_wP_S(v\Sigma^*)$ for any $w\in Q$\}\}}
\smallskip \hrule \smallskip

\begin{lemma}\label{lem:nosol}
  Let $P$ be a stochastic language and let $u_0, u_1, \ldots, u_n \in Res(P)$ be such that $\{u_0^{-1}P, u_1^{-1}P, \ldots,
  u_n^{-1}P\}$ is linearly independent.
 Then, with probability one, for any complete presentation $S$ of $P$,
 there exist a
positive number $\epsilon$ and an integer
$M$ such that $I(\{u_1, \ldots, u_n\},
u_0,S_m,\epsilon)$ has no solution for every $m\geq M$.
\end{lemma}

\begin{proof}
  Let $S$ be a complete presentation of $P$. Suppose that for every
  $\epsilon>0 $ and every integer $M$, there exists $m\geq M$ such that
  $I(\{u_1, \ldots, u_n\}, u_0,S_m,\epsilon)$ has a solution.  Then,
  for any integer $k$, there exists $m_k\geq k$ such that
  $I(\{u_1, \ldots, u_n\}, u_0,S_{m_k},1/k)$ has a solution
  $(\alpha_{1,k}, \ldots, \alpha_{n,k})$. 
Let $\rho_k=Max \{1, |\alpha_{1,k}|, \ldots, |\alpha_{n,k}|\}$, $\gamma_{0,k}=1/\rho_k$ and $\gamma_{i,k}=-\alpha_{i,k}/\rho_k$ for
$1\leq i \leq n$. For every $k$, $Max \{|\gamma_{i,k}|: 0\leq i\leq
n\}= 1$. Check that $$\forall k\geq 0, \left|\sum_{i=0}^n\gamma_{i,k}u_i^{-1}P_{S_{m_k}}(w\Sigma^*)\right|\leq
\frac{1}{\rho_kk}\leq \frac{1}{k}.$$

There exists a subsequence $(\alpha_{1,{\phi(k)}}, \ldots,
\alpha_{n,{\phi(k)}})$ of $(\alpha_{1,k}, \ldots, \alpha_{n,k})$ such
that \\$(\gamma_{0,{\phi(k)}}, \ldots, \gamma_{n,{\phi(k)}})$ converges
to $(\gamma_0, \ldots, \gamma_n)$. We show below that we should have
$\sum_{i=0}^n\gamma_{i}u_i^{-1}P(w\Sigma^*)=0$ for every word $w$,
which is contradictory with the independance assumption since $Max
\{\gamma_i: 0\leq i \leq n\}=1$. 

Let $w\in fact(supp(P))$. With
  probability 1, there exists an integer $k_0$ such that $w\in
  fact(S_{m_k})$ for any $k\geq k_0$. For such a $k$, we can write
$$\gamma_{i}u_i^{-1}P=(\gamma_{i}u_i^{-1}P-\gamma_{i}u_i^{-1}P_{S_{m_k}})+(\gamma_{i}-\gamma_{i,\phi(k)})u_i^{-1}P_{S_{m_k}}+\gamma_{i,\phi(k)}u_i^{-1}P_{S_{m_k}}$$
and therefore
$$\left|\sum_{i=0}^n\gamma_{i}u_i^{-1}P(w\Sigma^*)\right|\leq \sum_{i=0}^n|u_i^{-1}(P-P_{S_{m_k}})(w\Sigma^*))| + \sum_{i=0}^n|\gamma_{i}-\gamma_{i,\phi(k)}|+\frac{1}{k}$$
  which converges to 0 when $k$ tends to infinity.
\qed
\end{proof}

Let $P$ be a stochastic language over $\Sigma$, let
$\mathcal{A}=(A_{i})_{i\in I}$ be a family of subsets of $\Sigma^{*}$,
let $S$ be a finite sample drawn according to $P$, and let $P_{S}$ be
the empirical distribution associated with $S$. It can be shown
\cite{Vapnik98,Lugosi02} that for any confidence parameter $\delta$,
with a probability greater than $1-\delta$, for any $i\in
I$,\begin{equation} \left|P_{S}(A_{i})-P(A_{i})\right|\leq
c{\textstyle
\sqrt{\frac{\mathrm{VC}(\mathcal{A})-\log\frac{\delta}{4}}{Card(S)}}}\label{eq:vapnik}\end{equation}
where $\mathrm{VC}(\mathcal{A})$ is the dimension of
Vapnik-Chervonenkis of $\mathcal{A}$ and $c$ is a constant. 

When $\mathcal{A}=(\{ w\Sigma^{*}\})_{w\in\Sigma^{*}}$,
$\mathrm{VC}(\mathcal{A})\leq 2$. Indeed, let $r,s,t\in\Sigma^{*}$ and
let $Y=\{ r,s,t\}$. Let $u_{rs}$ (resp. $u_{rt},u_{st}$) be the longest
prefix shared by $r$ and $s$ (resp. $r$ and $t$, $s$ and $t$). One of
these 3 words is a prefix of the two other ones. Suppose that $u_{rs}$
is a prefix of $u_{rt}$ and $u_{st}$. Then, there exists no word $w$
such that $w\Sigma^{*}\cap Y=\{ r,s\}$. Therefore, no subset
containing more than two elements can be shattered by ${\mathcal{A}}$.

Let $\Psi(\epsilon,\delta)=\frac{c^{2}}{\epsilon^{2}}(2-\log\frac{\delta}{4})$. 

\begin{lemma}
\label{lem:Psi1}Let $P\in {\cal S}(\Sigma)$ and let $S$ be a complete presentation
of $P$. For any precision parameter $\epsilon$, any confidence parameter
$\delta$, any $n\geq\Psi\left(\epsilon,\delta\right)$, with a probability
greater than $1-\delta$, $\textrm{}\left|P_{n}(w\Sigma^*)-P(w\Sigma^*)\right|\leq\epsilon$
for all $w\in\Sigma^*$.
\end{lemma}

\begin{proof}
Use inequality (\ref{eq:vapnik}).  \qed
\end{proof}

Check that for any $\alpha$ such that $-1/2<\alpha<0$ and any
$\beta<-1$, if we define $\epsilon_k=k^{\alpha}$ and
$\delta_k=k^{\beta}$, there exists $K$ such that for all
$k\geq K$, we have $k\geq \Psi(\epsilon_k,\delta_k)$. For such choices
of $\alpha$ and $\beta$, we have $\lim_{k\rightarrow
\infty}\epsilon_k=0$ and $\sum_{k\geq 1}\delta_k<\infty$. 

\begin{lemma}\label{lem:existsol}
  Let $P\in {\cal S}(\Sigma)$, $u_0, u_1, \ldots, u_n \in
  res(P)$ and $\alpha_1, \ldots, \alpha_n\in {\mathbb R}$ be such that
  $u_0^{-1}P=\sum_{i=1}^n \alpha_iu_i^{-1}P$.  Then, with probability one, for any complete
  presentation $S$ of $P$, there exists $K$ s.t. $I(\{u_1, \ldots, u_n\},
  u_0,S_k,k^{-1/3})$ has a solution for every $k\geq K$.
\end{lemma}

\begin{proof}
  Let $S$ be a complete presentation of $P$. Let $\alpha_0=1$ and let
  $R=Max\{|\alpha_i|: 0\leq i\leq n\}$. With probability one, there
  exists $K_1$ s.t. $\forall k\geq K_1, \forall i=0,
  \ldots, n$, $|u_i^{-1}S_k|\geq
  \Psi([k^{1/3}(n+1)R]^{-1},[(n+1)k^2]^{-1})$. Let $k\geq K_1$. For
  any $X\subseteq \Sigma^*$, 
\small
$$|u_0^{-1}P_{S_k}(X)-\sum_{i=1}^n
\alpha_iu_i^{-1}P_{S_k}(X)|\leq
|u_0^{-1}P_{S_k}(X)-u_0^{-1}P(X)| +
\sum_{i=1}^n|\alpha_i||u_i^{-1}P_{S_k}(X)-u_i^{-1}P(X)|.$$
\normalsize
 From
Lemma~\ref{lem:Psi1}, with probability greater than
$1-1/k^2$, for any $i=0, \ldots, n$ and any word $w$,
$|u_i^{-1}P_{S_k}(w\Sigma^*)-u_i^{-1}P(w\Sigma^*)|\leq
[k^{1/3}(n+1)R]^{-1}$ and therefore,
$|u_0^{-1}P_{S_k}(w\Sigma^*)-\sum_{i=1}^n
\alpha_iu_i^{-1}P_{S_k}(w\Sigma^*)|\leq k^{-1/3}$.

For any integer $k\geq K_1$, let $A_k$ be the event:
$|u_0^{-1}P_{S_k}(w\Sigma^*)-\sum_{i=1}^n
\alpha_iu_i^{-1}P_{S_k}(w\Sigma^*)|> k^{-1/3}$. Since $Pr(A_k)<1/k^2$,
the probability that a finite number of $A_k$ occurs is 1.

Therefore, with probability 1, there exists an integer $K$ such that for any $k\geq K$, $I(\{u_1, \ldots, u_n\},
  u_0,S_k,k^{-1/3})$ has a solution. \qed
\end{proof}

\begin{lemma}\label{lem:paramconverg}
Let $P\in {\cal S}(\Sigma)$, let $u_0, u_1, \ldots, u_n \in
  res(P)$ such that $\{u_1^{-1}P, \ldots, u_n^{-1}P\}$ is linearly
  independent and let $\alpha_1, \ldots, \alpha_n\in {\mathbb R}$ be
  such that $u_0^{-1}P=\sum_{i=1}^n \alpha_iu_i^{-1}P$.  Then, with
  probability one, for any complete presentation $S$ of $P$, there
  exists an integer $K$ such that $\forall k\geq K$, any solution
  $\widehat{\alpha_1}, \ldots, \widehat{\alpha_n}$ of $I(\{u_1,
  \ldots, u_n\}, u_0,S_k,k^{-1/3})$ satisfies
  $|\widehat{\alpha_i}-\alpha_i|<O(k^{-1/3})$ for $1\leq i \leq n$.
\end{lemma}
\begin{proof}
  Let $w_1,\ldots, w_n\in \Sigma^*$ be such that the square matrix $M$
  defined by $M[i,j]=u_j^{-1}P(w_i\Sigma^*)$ for $1\leq i,j\leq n$ is
  inversible. Let $A=(\alpha_1, \ldots, \alpha_n)^t$,
  $U_0=(u_0^{-1}P(w_1\Sigma^*),\\ \ldots, u_0^{-1}P(w_n\Sigma^*))^t$. We
  have $MA=U_0$. Let $S$ be a complete presentation of $P$, let $k\in
  {\mathbb N}$ and let $\widehat{\alpha_1}, \ldots,
  \widehat{\alpha_n}$ be a solution of $I(\{u_1, \ldots, u_n\},
  u_0,S_k,k^{-1/3})$.  Let $M_k$ be the square matrix defined by
  $M_k[i,j]=u_j^{-1}P_{S_k}(w_i\Sigma^*)$ for $1\leq
  i,j\leq n$, let $A_k=(\widehat{\alpha_1}, \ldots,
  \widehat{\alpha_n})^t$ and $U_{0,k}=(u_0^{-1}P_{S_k}(w_1\Sigma^*),
  \ldots, u_0^{-1}P_{S_k}(w_n\Sigma^*))^t$. We have $$\Vert
  M_kA_k-U_{0,k}\Vert^2
  =\sum_{i=1}^n[u_0^{-1}P_{S_k}(w_i\Sigma^*)-\sum_{j=1}^n
    \widehat{\alpha_j}u_j^{-1}P_{S_k}(w_i\Sigma^*)]^2\leq
  nk^{-2/3}.$$
  Check that 
  \begin{center}
    $ A-A_k= M^{-1}(MA-U_0+U_0-U_{0,k}+U_{0,k}-M_kA_k+M_kA_k-MA_k)$
\end{center}
  and
  therefore, for any $ 1\leq i \leq n$
$$|\alpha_i-\widehat{\alpha_i}|\leq \Vert A-A_k\Vert \leq
\Vert M^{-1}\Vert (\Vert U_0-U_{0,k}\Vert +n^{1/2}k^{-1/3}+\Vert M_k-M\Vert \Vert A_k\Vert .$$  Now, by using Lemma~\ref{lem:Psi1}
and Borel-Cantelli Lemma as in the proof of Lemma~\ref{lem:existsol},
with probability 1, there exists $K$ such that for all $k\geq
K$, $\Vert U_0-U_{0,k}\Vert <O(k^{-1/3})$ and $\Vert M_k-M\Vert <O(k^{-1/3})$. Therefore, for
all $k\geq K$, any solution
$\widehat{\alpha_1}, \ldots, \widehat{\alpha_n}$ of $I(\{u_1, \ldots,
u_n\}, u_0,S_k,k^{-1/3})$ satisfies
$|\widehat{\alpha_i}-\alpha_i|<O(k^{-1/3})$ for $1\leq i \leq n$. \qed
\end{proof}

\begin{theorem}
Let $P\in {\cal S}_{\mathbb R}^{rat}(\Sigma)$ and $A$ be the prefixial reduced
representation of $P$. Then, with probability one, for any complete presentation
$S$ of $P$, there exists an integer $K$ such that for any
$k\geq K$, $DEES(S_k)$ returns a multiplicity automaton $A_k$ whose
support is the same as $A$'s. Moreover, there exists a constant $C$
such that for any parameter
$\alpha$ of $A$, the corresponding parameter $\alpha_k$ in $A_k$
satisfies $|\alpha-\alpha_k|\leq Ck^{-1/3}$.
\end{theorem}
\begin{proof}
  Let $Q_P$ be the set of states of $A$, i.e. the smallest prefixial
  subset of $res(P)$ such that $\{u^{-1}P:u\in Q_P\}$ spans the same
  vector space as $Res(P)$. Let $u\in Q_P$, let $Q_u=\{v\in Q_P|
  v<u\}$ and let $x\in \Sigma$.
  \begin{itemize}
  \item If $\{v^{-1}P|v\in Q_u\cup\{ux\}\}$ is linearly independent,
    from Lemma~\ref{lem:nosol}, with probability 1, there exists
    $\epsilon_{ux}$ and $K_{ux}$ such that for any $k\geq K_{ux}$, $I(Q_u,
    ux,S_k,\epsilon_{ux})$ has no solution.
  \item If there exists $(\alpha_v)_{v\in Q_u}$ such that
        $(ux)^{-1}P=\sum_{v\in Q_u}\alpha_vv^{-1}P$, from
        Lemma~\ref{lem:existsol}, with probability 1, there exists an
        integer $K_{ux}$ such that for any $k\geq K_{ux}$, $I(Q_u, ux,S_k,k^{-1/3})$ has a solution.
  \end{itemize}
Therefore, with probability one, there exists an integer $K$ such that
for any $k\geq K$,  $DEES(S_k)$
returns a multiplicity automaton $A_k$ whose set of states is equal to
$Q_P$. Use
Lemmas~\ref{lem:Psi1}~and~\ref{lem:paramconverg} to check the last part
of the proposition.\qed
\end{proof}

When the target is in ${\cal S}_{\mathbb Q}^{rat}(\Sigma)$, DEES can
be used to exactly identify it. The proof is based on the
representation of real numbers by continuous fraction. See
\cite{HardyWright79} for a survey on continuous fraction and
\cite{DenisEsposito04} for a similar application.

Let $(\epsilon_{n})$ be a sequence of non negative real numbers which
converges to $0$, let $x\in{\mathbb Q}$, let $(y_{n})$ be a sequence
of elements of ${\mathbb Q}$ such that $|x-y_{n}|\leq\epsilon_{n}$ for
all but finitely many $n$. It can be shown that there exists an
integer $N$ such that, for any $n\geq N$, $x$ is the unique rational
number $\frac{p}{q}$ which satisfies
$\left|y_{n}-\frac{p}{q}\right|\leq\epsilon_{n}\leq\frac{1}{q^{2}}$.
Moreover, the unique solution of these inequations can be computed
from $y_n$. 

Let $P\in {\cal S}_{\mathbb Q}^{rat}(\Sigma)$, let $S$ be a complete
presentation of $P$ and let $A_k$ the MA output by DEES on input
$S_k$. Let $\overline{A}_k$ be the MA derived from $A_k$ by replacing
every parameter $\alpha_k$ with a solution $\frac{p}{q}$ of
$\left|\alpha-\frac{p}{q}\right|\leq k^{-1/4} \leq\frac{1}{q^{2}}$.

\begin{theorem}
Let $P\in {\cal S}_{\mathbb Q}^{rat}(\Sigma)$ and $A$ be the prefixial reduced
representation of $P$. Then, with probability one, for any complete presentation
$S$ of $P$, there exists an integer $K$ such that
$\forall k\geq K$, $DEES(S_k)$ returns an MA $A_k$ such
that $\overline{A}_k=A$.
\end{theorem}
\begin{proof}From previous theorem, for every parameter
  $\alpha$ of $A$, the corresponding parameter $\alpha_k$ in $A_k$
  satisfies $|\alpha-\alpha_k|\leq Ck^{-1/3}$ for some constant $C$.
  Therefore, if $k$ is sufficiently large, we have
  $|\alpha-\alpha_k|\leq k^{-1/4}$ and there exists an integer $K$
  such that $\alpha=p/q$ is the unique solution of $\left|\alpha-\frac{p}{q}\right|\leq k^{-1/4} \leq\frac{1}{q^{2}}$.\qed
\end{proof}

\section{Learning rational stochastic languages}\label{lrsl}

We have seen that ${\cal S}_{\mathbb Q}^{rat}(\Sigma)$ is identifiable
in the limit. Moreover, DEES runs in polynomial time and aims at
computing a representation of the target which is minimal and whose
parameters depends only on the target to be learned. DEES computes
estimates which are proved to converge reasonably fast to these
parameters. That is, DEES compute functions which are likely to be
close to the target. But these functions are not stochastic languages
and it remains to study how they can be used in a grammatical
inference perspective. 

Any rational stochastic language $P$ defines a vector subspace of
${\mathbb R}\langle\langle\Sigma\rangle\rangle$ in which the
stochastic languages form a compact convex subset. 

\begin{proposition}\label{compactsto}
  Let $p_1, \ldots, p_n$ be $n$ independent stochastic languages.
  Then, $\Lambda=\{\overrightarrow{\alpha}=$ $(\alpha_1, \ldots,
  \alpha_n)\in {\mathbb R}^n: \sum_{i=1}^n\alpha_ip_i\in {\cal
    S}(\Sigma)\}$ is a compact convex subset of ${\mathbb R}^n$.
\end{proposition}
\begin{proof}
  First, check that for any
  $\overrightarrow{\alpha}, \overrightarrow{\beta}\in \Lambda$ and any
  $\gamma \in [0,1]$, the series $\sum_{i=1}^{n}[\gamma
  \alpha_i+(1-\gamma)\beta_i]p_i$ is a stochastic language. Hence,
  $\Lambda$ is convex.

For every word $w$, the mapping $\overrightarrow{\alpha}
\rightarrow \sum_{i=1}^n\alpha_ip_i(w)$ defined from ${\mathbb R}^n$
into ${\mathbb R}$ is linear; and so is the mapping $\overrightarrow{\alpha} \rightarrow \sum_{i=1}^n\alpha_i$. $\Lambda$ is closed since these mappings are
  continuous and since $$\Lambda=\left\{\overrightarrow{\alpha}\in {\mathbb
    R}^n: \sum_{i=1}^n\alpha_ip_i(w)\geq 0\textrm{ for every word }
    w\textrm{ and} \sum_{i=1}^n\alpha_i=1\right\}.$$ 
  
  Now, let us show that $\Lambda$ is bounded. Suppose that for any
  integer $k$, there exists $\overrightarrow{\alpha}_k \in \Lambda$
  such that $\Vert \overrightarrow{\alpha}_k\Vert \geq k$. Since
  $\overrightarrow{\alpha}_k/\Vert \overrightarrow{\alpha}_k\Vert$
  belongs to the unit sphere in ${\mathbb R}^n$, which is compact,
  there exists a subsequence $\overrightarrow{\alpha}_{\phi(k)}$ such
  that $\overrightarrow{\alpha}_{\phi(k)}/\Vert
  \overrightarrow{\alpha}_{\phi(k)}\Vert$ converges to some
  $\overrightarrow{\alpha}$ satisfying $\Vert
  \overrightarrow{\alpha}\Vert=1$. Let
  $q_k=\sum_{i=1}^n\alpha_{i,k}p_i$ and $r=\sum_{i=1}^n\alpha_ip_i$.

  For any $0<\lambda\leq \Vert \overrightarrow{\alpha}_k\Vert$,
  $p_1+\lambda\frac{q_{k}-p_1}{\Vert
    \overrightarrow{\alpha}_k\Vert}=(1-\frac{\lambda}{\Vert
    \overrightarrow{\alpha}_k\Vert})p_1+\frac{\lambda}{\Vert
    \overrightarrow{\alpha}_k\Vert}q_{k}$ is a stochastic language
  since ${\cal S}(\Sigma)$ is convex; for every $\lambda>0$,
  $p_1+\lambda\frac{q_{\phi(k)}-p_1}{\Vert
  \overrightarrow{\alpha}_{\phi(k)}\Vert}$ converges to
  $p_1+\lambda r$ when $l\rightarrow \infty$, which is a stochastic
  language since $\Lambda$ is closed. Therefore, for any $\lambda>0$,
  $p_1+\lambda r$ is a stochastic language. Since $p_1(w)+\lambda
  r(w)\in [0,1]$ for every word $w$, we must have $r=0$, i.e.
  $\alpha_i=0$ for any $1\leq i \leq n$ since the languages $p_1,
  \ldots, p_n$ are independent, which is impossible since
  $\Vert
  \overrightarrow{\alpha}\Vert=1$. Therefore, $\Lambda$ is bounded.  \qed
\end{proof}

The MA $A$ output by DEES generally do not compute stochastic
languages. However, we wish that the series $r_A$ they compute share
some properties with them. Next proposition gives sufficient
conditions which guaranty that $\sum_{k\geq 0}r_A(\Sigma^k)=1$.

\begin{proposition}\label{sumbounded}
  Let $A=\left\langle \Sigma,Q=\{q_1, \ldots,
    q_n\},\varphi,\iota,\tau\right\rangle $ be an MA and let $M$
  be the square matrix defined by
  $M[i,j]=\left[\varphi(q_i,\Sigma,q_j)\right]_{1\leq i,j\leq n}$.
  Suppose that the spectral radius of $M$ satisfies $\rho(M)<1$. Let
  $\overrightarrow{\iota}=(\iota(q_1), \ldots, \iota(q_n))$ and
  $\overrightarrow{\tau}=(\tau(q_1), \ldots, \tau(q_n))^t$.
\begin{enumerate}
\item Then, the matrix $(I-M)$ is inversible and
    $\sum_{k\geq 0}M^k$ converges to $(I-M)^{-1}$. 
  \item $\forall q_i\in Q, \forall K\geq 0$, $\sum_{k\geq
      K}r_{A,q_i}(\Sigma^k)$ converges to
    $M^K\sum_{j=1}^n(I-M)^{-1}[i,j]\tau(q_j)$ and
    $\sum_{k\geq K}r_{A}(\Sigma^k)$ converges to
    $\overrightarrow{\iota}M^K(I-M)^{-1}\overrightarrow{\tau}$.
  \item If $\forall q\in Q, \tau(q)+\varphi(q,\Sigma,Q)=1$, then
    $\forall q\in Q, r_{A,q}(\sum_{k\geq 0}\Sigma^k)=1$. If moreover
    $\sum_{q\in Q}\iota(q)=1$, then $r(\sum_{k\geq 0}\Sigma^k)=1$.
\end{enumerate}
\end{proposition}

\begin{proof}\begin{enumerate}
\item Since $\rho(M)<1$, 1 is not an eigen value of $M$ and $I-M$ is
    inversible. From Gelfand's formula,
    $\lim_{k\rightarrow\infty}\Vert M^k\Vert =0$. Since for any integer $k$,
    $(I-M)(I+M+\ldots +M^k)=I-M^{k+1}$, the sum $\sum_{k\geq 0}M^k$
    converges to $(I-M)^{-1}$.
\item Since $r_{A,q_i}(\Sigma^k)=\sum_{j=1}^nM^k[i,j]\tau(q_j)$,
$\sum_{k\geq K} r_{A,q_i}(\Sigma^k)=M^K\sum_{j=1}^n
(1-M)^{-1}[i,j]\tau(q_j)$ and $\sum_{k\geq
K}r_{A}(\Sigma^k)=\sum_{i=1}^n\iota(q_i)r_{A,q_i}(\Sigma^{\geq
K})=\overrightarrow{\iota}M^K(I-M)^{-1}\overrightarrow{\tau}$.
\item Let $s_i=r_{A,q_i}(\Sigma^*)$ for $1\leq i \leq n$ and
  $\overrightarrow{s}=(s_1, \ldots, s_n)^t$. We have
  $(I-M)\overrightarrow{s}=\overrightarrow{\tau}$. Since $I-M$ is
  inversible, there exists one and only one $s$ such that
  $(I-M)\overrightarrow{s}=\overrightarrow{\tau}$. But since
  $\tau(q)+\varphi(q,\Sigma,Q)=1$ for any state $q$, the vector
  $(1, \ldots, 1)^t$ is clearly a solution. Therefore, $s_i=1$ for
  $1\leq i \leq n$.  If $\sum_{q\in Q}\iota(q)=1$, then
  $r(\Sigma^*)=\sum_{q\in Q}\iota(q)r_{A,q}(\Sigma^*)=1$.\qed
\end{enumerate}
\end{proof}


\begin{proposition}\label{spectralradiusbounded}
  Let $A=\left\langle \Sigma,Q,\varphi,\iota,\tau\right\rangle $ be
  a reduced representation of a stochastic language
  $P$.  Let $Q=\{q_1, \ldots,q_n\}$ and let $M$ be the square matrix
  defined by $M[i,j]=\left[\varphi(q_i,\Sigma,q_j)\right]_{1\leq
    i,j\leq n}$.  Then the spectral radius of $M$ satisfies
  $\rho(M)<1$.
\end{proposition}
\begin{proof} 
  From Prop.~\ref{compactsto}, let $R$ be such that
  $\{\overrightarrow{\alpha}\in {\mathbb R}^n:
  \sum_{i=1}^n\alpha_iP_{A,q_i}\in {\cal S}(\Sigma)\}\subseteq
  B(0,R)$.  For every $u\in res(P_A)$ and every $1\leq i\leq n$, we have
  $$u^{-1}P_{A,q_i}=\frac{\sum_{1\leq j\leq
      n}\varphi(q_i,u,q_j)P_{A,q_j}}{P_{A,q_i}(u\Sigma^*)}\cdot$$
  Therefore, for every word $u$ and every $k$, we have
  $\left|\varphi(q_i,u,q_j)\right|\leq R\cdot P_{A,q_i}(u\Sigma^*)$
  and
$$\left|\varphi(q_i,\Sigma^k,q_j)\right|\leq \sum_{u\in \Sigma^k}\left|\varphi(q_i,u,q_j)\right|\leq R\cdot
P_{A,q_i}(\Sigma^{\geq k}).$$

Now, let $\lambda$ be an
eigen value of $M$ associated with the eigen vector $v$ and let $i$ be
an index such that $|v_i|=Max \{|v_j|: j=1, \ldots, n\}$. For every
integer $k$, we have
$$M^kv=\lambda^kv\textrm{ and }|\lambda^k v_i|=
|\sum_{j=1}^n\varphi(q_i,\Sigma^k,q_j)v_j|\leq nR\cdot
P_{A,q_i}(\Sigma^{\geq k})|v_i|$$
which implies that $|\lambda|<1$
since $P_{A,q_i}(\Sigma^{\geq k})$ converges to 0 when $k\rightarrow \infty$. \qed
\end{proof}

If the spectral radius of a matrix is $<1$, the power of $M$ decrease
exponentially fast. 

\begin{lemma}\label{lem:Crhok}
Let $M\in {\mathbb R}^{n\times
      n}$ be such that $\rho(M)<1$. Then,  there exists $C\in {\mathbb
      R}$ and $\rho\in [0,1[$
      such that for any
      integer $k\geq 0$, $\Vert M^k\Vert \leq C\rho^k$.
\end{lemma}
\begin{proof}
  Let $\rho\in ]\rho(M),1[$. From Gelfand's formula, there exists an
  integer $K$ such that for any $k\geq K$, $\Vert M^k\Vert ^{1/k}\leq
  \rho$.  Let $C=Max\{\Vert M^h\Vert /\rho^h:h<K\}$. Let $k\in
  {\mathbb N}$ and let $a,b\in
  {\mathbb N}$ be such that $k=aK
  +b$ and $b<K$. We have $$\Vert M^k\Vert =\Vert M^{aK+b}\Vert \leq
  \Vert M^{aK}\Vert\Vert M^b\Vert \leq \rho^{aK}\Vert M^b\Vert \leq
  \rho^{k}\frac{\Vert M^b\Vert }{\rho^b} \leq C\rho^{k}.$$
\end{proof}

\begin{proposition}
  Let $P\in {\cal S}^{rat}_{\mathbb R}(\Sigma)$. There exists a
  constant $C$ and $\rho\in [0,1[$ such that for any integer $k$,
  $P(\Sigma^{\geq k})\leq C\rho^k$. 
\end{proposition}
\begin{proof}
  Let $A=\left\langle \Sigma,Q,\varphi,\iota,\tau\right\rangle $ be a
  reduced representation of $P$ and let $M$ be the square matrix
  defined by $M[i,j]=\left[\varphi(q_i,\Sigma,q_j)\right]_{1\leq
    i,j\leq n}$.  From Prop.~\ref{spectralradiusbounded}, the spectral
  radius of $M$ is <1. From Lemma~\ref{lem:Crhok}, there exists $C_1$
  and $\rho\in [0,1[$ such that $\Vert M^k\Vert \leq C_1\rho^k$ for
  every integer $k$. Let $\overrightarrow{\iota_A}=(\iota(q_1),
  \ldots, \iota(q_n))$ and $\overrightarrow{\tau_A}=(\tau(q_1),
  \ldots, \tau(q_n))^t$. We have $$P(\Sigma^{\geq k}) \leq \Vert
  \iota_A\Vert \cdot \Vert M^{k}\Vert \cdot \Vert (I-M)^{-1}\Vert
  \cdot \Vert \overrightarrow{\tau_A}\Vert \leq C\rho^k$$
  with $C=
  C_1\Vert \overrightarrow{\iota_A}\Vert \cdot \Vert (1-M)^{-1}\Vert
  \cdot \Vert \overrightarrow{\tau_A}\Vert $.  
  \qed
\end{proof}

It is not difficult to design an MA $A$ which generates a stochastic
language $P$ and such that $\varphi(q,u,q')$ is unbounded when $u\in \Sigma^*$.
However, the next proposition proves that this situation never happens
when $A$ is a reduced representation of $P$.

\begin{proposition}
Let $P\in {\cal S}^{rat}_{\mathbb R}(\Sigma)$ and let $A=\left\langle
\Sigma,Q,\varphi,\iota,\tau\right\rangle $ be a reduced
representation of $P$.  Then, there exists a constant $C$ and $\rho\in
[0,1[$ such that for any integer $k$ and any pair of states
$q,q'$, $\sum_{u\in \Sigma^k}|\varphi(q,u,q')|\leq C\rho^k$.
\end{proposition}
\begin{proof}
  Let $k$ be an integer and let $q,q'\in Q$. Let $P_k=\{u\in \Sigma^k:
  \varphi(q,u,q')\geq 0\}$ and $N_k=\Sigma^k\setminus
  P_k$. 
$$P_k^{-1}P_{A,q}=\sum_{u\in
P_k}\frac{P_{A,q}(u\Sigma^*)}{\sum_{u\in
P_k}P_{A,q}(u\Sigma^*)}u^{-1}P_{A,q}=\sum_{q''\in Q}\frac{\sum_{u\in
P_k}\varphi(q,u,q'')}{\sum_{u\in P_k}P_{A,q}(u\Sigma^*)}P_{A,q''}$$ is a
stochastic language which is a linear combination of the
independent stochastic languages $P_{A,q''}$. From prop.~\ref{compactsto}, there
exists a constant $R$ which depends only on $A$ s.t. $$
\left|\sum_{u\in P_k}\varphi(q,u,q')\right|=\sum_{u\in
P_k}\varphi(q,u,q')\leq R\sum_{u\in P_k}P_{A,q}(u\Sigma^*).$$
Similarly, we have $ \left|\sum_{u\in
N_k}\varphi(q,u,q')\right|=\sum_{u\in N_k}|\varphi(q,u,q')|\leq
R\sum_{u\in N_k}P_{A,q}(u\Sigma^*).$ Let $C$ and $\rho\in ]0,1[$ be
such that $P_{A,q}(\Sigma^{\geq k})\leq C\rho^k$ for any state $q$ and
any integer $k$. We have $$\sum_{u\in
\Sigma^k}|\varphi(q,u,q')|\leq R\sum_{u\in
\Sigma^k}P_{A,q}(u\Sigma^*)\leq RC\rho^k.$$\qed
\end{proof}

\begin{figure}[htbphh]
  \centering
\epsfig{figure=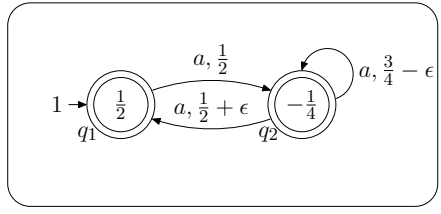, height=3cm} 
\caption{These MA compute a series $r_{\epsilon}$ such that
$\sum_{w\in \Sigma^*}r_{\epsilon}(w)=1$ if $\epsilon\neq 0$ and
$\sum_{w\in \Sigma^*}r_0(w)=2/5$. Note that when
$\epsilon=0$, the series $r_{0,q_1}$ and  $r_{0,q_2}$ are dependent.}
\label{fig:instable} 
\end{figure}

MA representation of rational stochastic languages are unstable (see Fig.~\ref{fig:instable}).
Arbitrarily close to an MA $A$ which generates a stochastic language,
we can find an MA $B$ such that the sum $\sum_{w\in \Sigma^*}r_B(w)$
converges to any real number or even diverges. However, the next
theorem shows that when $A$ is a reduced representation of a
stochastic language, any MA $B$ sufficiently close to $A$ defines a
series which is absolutely convergent. Moreover, simple syntactical
conditions ensure that $r_{B}(\Sigma^*)=1$.

\begin{theorem}
  Let $P\in {\cal S}^{rat}_{\mathbb R}(\Sigma)$ and let
  $A=\left\langle \Sigma,Q,\varphi_A,\iota_A,\tau_A\right\rangle $ be a
  reduced representation of $P$. Let $C_A$ and $\rho_A\in
  ]0,1[$ be such that for any integer $k$ and any pair of states
  $q,q'$, $\sum_{u\in \Sigma^k}|\varphi_A(q,u,q')|\leq
  C_A\rho_A^k$. Then, for any $\rho>\rho_A$, there exists $C$ and
  $\alpha>0$ such that for any MA $B=\left\langle
    \Sigma,Q,\varphi_B,\iota_B,\tau_B\right\rangle $ satisfying
  \begin{equation}
    \label{eq:1}
    \forall q,q'\in Q, \forall x\in \Sigma, |\varphi_A(q,x,q')-\varphi_B(q,x,q')|<\alpha
  \end{equation} we have $\sum_{u\in \Sigma^k}|\varphi_B(q,u,q')|\leq C\rho^k$
  for any pair of states $q,q'$ and any integer $k$. As a consequence,
  the series $r_B$ is absolutely convergent. Moreover, if $B$
  satisfies also
  \begin{equation}
\forall q\in Q, \tau_B(q)+\varphi_B(q,\Sigma,Q)=1\textrm{ and }\sum_{q\in
    Q}\iota_B(q)=1\label{eq:2}
\end{equation}
then, $\alpha$ can be chosen such that (\ref{eq:1}) implies that $r_{B,q}(\Sigma^*)=1$ for any state $q$ and $r_{B}(\Sigma^*)=1$.
\end{theorem}
\begin{proof}
Let $k$ be such that $(2nC_A)^{1/k}\leq \rho/\rho_A$ where
  $n=|Q|$. There exists $\alpha>0$ such that for any MA
  $B=\left\langle \Sigma,Q,\varphi_B,\iota_B,\tau_B\right\rangle $
  satisfying~(\ref{eq:1}), we have $$\forall q,q'\in
Q, \sum_{u\in
  \Sigma^k}|\varphi_B(q,u,q')-\varphi_A(q,u,q')|<C_A\rho_A^k.$$ 

Since $\sum_{u\in
  \Sigma^k}|\varphi_A(q,u,q')|\leq C_A\rho_A^k$, we must have also
$$\sum_{u\in
  \Sigma^k}|\varphi_B(q,u,q')|\leq 2C_A\rho_A^k\leq \frac{\rho^k}{n}\cdot$$

Let $C_1= Max \{\sum_{u\in \Sigma^{<k}}|\varphi_B(q,u,q')|: q,q'\in
Q\}$. Let $l,a,b\in {\mathbb N}$ such that
$l=ak+b$ and $b<k$. Let $u\in \Sigma^l$ and let $u=u_0\ldots u_a$
where $|u_i|=k$ for $0\leq i< a$ and $|u_a|=b$. For any pair of
states $q_0,q_{a+1}$, we have
$$\varphi_B(q_0,u,q_{a+1})=\sum_{q_1, \ldots, q_a\in Q}\prod_{i=0}^a
\varphi_B(q_i,u_i,q_{i+1})$$
and 
\begin{align*}
\sum_{u\in \Sigma^l}\varphi_B(q_0,u,q_{a+1})&=\sum_{u_0,\ldots, u_{a-1}\in \Sigma^k}\sum_{u_a\in \Sigma^b}\sum_{q_1, \ldots, q_a\in Q}\prod_{i=0}^a
\varphi_B(q_i,u_i,q_{i+1})\\
&=\sum_{q_1, \ldots, q_a\in Q}\sum_{u_0,\ldots, u_{a-1}\in \Sigma^k}\sum_{u_a\in \Sigma^b}\prod_{i=0}^a
\varphi_B(q_i,u_i,q_{i+1})\\
&=\sum_{q_1, \ldots, q_a\in Q}\prod_{i=0}^{a-1}\left(\sum_{u\in \Sigma^k}\varphi_B(q_i,u,q_{i+1})\right)\left(\sum_{u\in \Sigma^b}
\varphi_B(q_a,u_i,q_{a+1})\right).\\
\end{align*}
Hence, 
$\sum_{u\in \Sigma^l}|\varphi_B(q_0,u,q_{m+1})|\leq n^a\cdot
\left(\frac{\rho^k}{n}\right)^a\cdot C_1\leq C\rho^l$ where $C=\frac{C_1}{\rho^{k-1}}$.

Now, let us prove that $r_B$ is absolutely convergent. $$\sum_{w\in
\Sigma^*}|r_B(w)|\leq \sum_{k\in {\mathbb N}}\sum_{u\in
\Sigma^k}\sum_{q,q'\in Q}\iota_B(q)\varphi_B(q,u,q')\tau_B(q')\leq C'$$
where $C'=Cn^2Max\{|\iota_B(q)\tau_B(q')|: q,q'\in Q\}/(1-\rho)$.

Lastly, let $M_B$ be the square matrix defined by
$M_B[i,j]=\varphi_B(q_i,\Sigma,q_j)$. Since the spectral radius of a
matrix depends continuously on its coefficients and since $A$ is a
reduced representation of a stochastic language, any MA
satisfying~(\ref{eq:1}) with $\alpha$ sufficiently small must have a
spectral radius
<1~(Prop.~\ref{spectralradiusbounded}). Therefore, if $B$
satisfies~(\ref{eq:2}) and~(\ref{eq:1}) with $\alpha$ sufficiently
small, the Prop.~\ref{sumbounded} entails
the conclusion.\qed
\end{proof}

It remains to show how a series which converges absolutely to 1 can be
used to approximate a stochastic language. 

Let $r$ be a series over $\Sigma$ such that $\sum_{w\in
\Sigma^*}r(w)$ converges absolutely to 1. Therefore, $r(X)=\sum_{u\in
X}r(u)$ is defined without ambiguity for every $X\subseteq \Sigma^*$
and $r(X)$ is bounded by $\overline{r}= \sum_{u\in \Sigma^*}|r(u)|$. Let $S$ be the
smallest subset of $\Sigma^*$ such that $$\varepsilon\in S\textrm{ and
} \forall u\in \Sigma^*, \forall x\in \Sigma, u\in S \textrm{ and
}r(ux\Sigma^*)>0 \Rightarrow ux\in S.$$ $S$ is a prefixial subset of
$\Sigma^*$ and $\forall u\in S, r(u\Sigma^*)>0$.  For every word $u\in
S$, let us define $N(u)=\cup\{ux\Sigma^*: x\in \Sigma,
r(ux\Sigma^*)\leq 0\}\cup \{u: \textrm{ if }r(u)\leq 0\}\textrm{ and
}N=\cup\{N(u): u\in \Sigma^*\}.$ Then, for every $u\in S$, let us define
$\lambda_u$ by:
$$\lambda_{\varepsilon}=(1-r(N(\varepsilon)))^{-1} \textrm{ and }
\lambda_{ux}=\lambda_u\frac{r(ux\Sigma^*)}{r(ux\Sigma^*)-r(N(ux))}.$$

\begin{lemma}\label{lem:lambda}
For every word $u\in S$, $e^{r(N)/\overline{r}}\leq \lambda_u\leq 1$.
\end{lemma}
\begin{proof}First, check that $r(N(u))\leq 0$ for every $u\in S$. Therefore,  $\lambda_u\leq 1$. Now, check that 
  if $u,uv\in S$ then $v=\varepsilon$ or $N(u)\cap
  N(uv)=\emptyset$. Let $u=x_1\ldots x_n\in \Sigma^*$ where
  $x_1,\ldots, x_n\in \Sigma$ and let $u_0=\epsilon$ and
  $u_i=u_{i-1}x_i$ for $1\leq i \leq n$.  We have
  $$\lambda_u=\prod_{i=0}^n\frac{r(u_i\Sigma^*)}{r(u_i\Sigma^*)-r(N(u_i))}=\prod_{i=0}^n\left(1-\frac{r(N(u_i))}{r(u_i\Sigma^*)}\right)^{-1}$$
  and $$\log{\lambda_u}=-\sum_{i=0}^n\log{\left(1-\frac{r(N(u_i))}{r(u_i\Sigma^*)}\right)}\geq
  \sum_{i=0}^n\frac{r(N(u_i))}{r(u_i\Sigma^*)}\cdot$$
Since $r(u_i\Sigma^*)\leq \overline{r}$, 
$\log{\lambda_u}\geq \sum_{i=0}^nr(N(u_i))/\overline{r} =r(\cup_{i=0}^nN(u_i))/\overline{r}\geq r(N)/\overline{r}.$
  Therefore,
  $\lambda_u \geq e^{r(N)/\overline{r}}$.\qed
\end{proof}

Let $p_r$ be the
series defined by: $p_r(u)=0$ if $u\in N$ and
$p_r(u)=\lambda_ur(u)$ otherwise. We show that $pr$ is a stochastic language.

\begin{lemma}\label{lem:pr}
  \begin{itemize}
  \item $p_r(\varepsilon)+\lambda_{\varepsilon}\sum_{x\in
      S\cap \Sigma}r(x\Sigma^*)=1$,
\item For any $u\in \Sigma^*$ and any $x\in \Sigma$, if $ux\in S$ then
$$p_r(ux)+\lambda_{ux}\sum_{\{y\in \Sigma: uxy\in S\}}r(uxy\Sigma^*)=\lambda_ur(ux\Sigma^*).$$
  \end{itemize}
\end{lemma}
\begin{proof}
First, check that for every $u\in S$, $$p_r(u)+\lambda_u\sum_{x\in u^{-1}S\cap\Sigma}r(ux\Sigma^*)=\lambda_u(r(u\Sigma^*)-r(N(u)).$$
Then, $p_r(\varepsilon)+\lambda_{\varepsilon}\sum_{x\in
      S\cap \Sigma}r(x\Sigma^*)=\lambda_{\varepsilon}(1-r(N(\varepsilon)))=1.$ Now, let  $u\in \Sigma^*$ and $x\in \Sigma$ s.t. $ux\in S$, 
$p_r(ux)+\lambda_{ux}\sum_{\{y\in \Sigma: uxy\in S\}}r(uxy\Sigma^*)=\lambda_{ux}(r(ux\Sigma^*)-r(N(ux)))=\lambda_ur(ux\Sigma^*).$\qed
\end{proof}
\begin{lemma}\label{lem:Qpref}Let $Q$ be a prefixial finite subset of $\Sigma^*$ and
let $Q_s=(Q\Sigma\setminus Q)\cap S$. Then
$$p_r(Q)=1-\sum_{ux\in Q_s, x\in \Sigma}\lambda_ur(ux\Sigma^*).$$
\end{lemma}
\begin{proof}
By induction on $Q$.  When $Q=\{\varepsilon\}$, the relation comes
directly from Lemma~\ref{lem:pr}.
Now, suppose that the relation is true for a prefixial subset $Q'$,
let $u_0\in Q'$ and $x_0\in \Sigma$ such that $u_0x_0\not\in Q'$ and let
$Q=Q'\cup\{u_0x_0\}$. We have $$p_r(Q)=p_r(Q')+p_r(u_0x_0)=1-\sum_{ux\in Q'_s, x\in \Sigma}\lambda_ur(ux\Sigma^*) + p_r(u_0x_0)$$ where
$Q'_s=(Q'\Sigma\setminus Q')\cap S$, from inductive hypothesis.

If $u_0x_0\not\in S$, check that $p_r(u_0x_0)=0$ and that $Q_s=Q'_s$. Therefore, $p_r(Q)=1-\sum_{ux\in Q_s, x\in \Sigma}\lambda_ur(ux\Sigma^*).$

If $u_0x_0\in S$, then $
Q_s=Q'_s\setminus\{u_0x_0\}\cup (u_0x_0\Sigma\cap S)$. Therefore,
\begin{align*}
p_r(Q)&=1-\sum_{ux\in Q'_s, x\in \Sigma}\lambda_ur(ux\Sigma^*) + p_r(u_0x_0)\\
&=1-\sum_{ux\in Q_s, x\in \Sigma}\lambda_ur(ux\Sigma^*)-\lambda_{u_0}r(u_0x_0\Sigma^*)\\
&+\lambda_{u_0x_0}\sum_{u_0x_0x\in S, x\in \Sigma}r(u_0x_0x\Sigma^*) + p_r(u_0x_0)\\
&=1-\sum_{ux\in Q_s, x\in \Sigma}\lambda_ur(ux\Sigma^*)\textrm{ from Lemma~\ref{lem:pr}.}\hspace*{4cm}\qed
\end{align*}
\end{proof}
\begin{proposition}
  Let $r$ be a formal series over $\Sigma$ such that $\sum_{w\in
    \Sigma^*}r(w)$ converges absolutely to 1. Then, $p_r$ is a
  stochastic language such that for every $u\in \Sigma^*\setminus N$,
  $$(1+r(N)/\overline{r})r(u)\leq e^{r(N)/\overline{r}}r(u) \leq p_r(u)\leq r(u).$$
\end{proposition}
\begin{proof} From Lemma~\ref{lem:lambda}, the only thing that
  remains to be proved is that $p_r$ is a stochastic language.
  Clearly, $p_r(u)\in [0,1]$ for every word $u$. From
  Lemma~\ref{lem:Qpref}, for any integer $k$, $$|1-p_r(\Sigma^{\leq
  k})| \leq \sum_{u\in \Sigma^{k+1}\cap S}r(u\Sigma^*)\leq r(\Sigma^{>
  k})$$ which tends to 0 since $r$ is absolutely convergent.\qed
\end{proof}

To sum up, DEES computes MA $A$ whose structure is equal to the
structure of the target from some steps, and whose parameters tends
reasonably fast to the true parameters. From some steps, they define
absolutely rational series $r_A$ which converge absolutely to 1. By
using these MA, it is possible to efficiently compute $p_{r_A}(u)$ or
$p_{r_A}(u\Sigma^*)$ for any word $u$. Moreover, since $r_A$
converges absolutely and since $A$ tends to the target, the weight
$r_A(N)$ of the negative values tends to 0 and $p_{r_A}$ converges to
the target. 









\section{Conclusion}\label{conclusion}
We have defined an inference algorithme DEES designed to learn
rational stochastic languages which strictly contains the class of
stochastic languages computable by PA (or HMM). We have shown that the
class of rational stochastic languages over ${\mathbb Q}$ is strongly
identifiable in the limit. Moreover, DEES is an efficient inference
algorithm which can be used in practical cases of grammatical
inference. The experiments we have already carried out confirm the
theoretical results of this paper: the fact that DEES aims at building
a natural and minimal representation of the target provides a very
significant improvement of the results obtained by classical
probabilistic inference algorithms. 
\bibliographystyle{plain}

\end{document}